\documentclass[10pt, a4paper, conference, compsocconf]{IEEEtran}

\usepackage{graphicx}
\usepackage{cite}

\ifCLASSINFOpdf

\else

\fi

\usepackage{url}
\usepackage[symbol]{footmisc}
\usepackage{color}
\usepackage{hyperref}
\usepackage{xcolor}
\hypersetup{
    colorlinks,
    linkcolor={red!50!black},
    citecolor={blue!50!black},
    urlcolor={blue!80!black}
}
\usepackage{subfig}
\usepackage{amsmath}
\usepackage{amsfonts}

\begin{document}

\title{A low-cost, flexible and portable volumetric capturing system}

\author{\IEEEauthorblockN{Vladimiros Sterzentsenko\thanks{Indicates equal contribution.}\thanks{vladster@iti.gr},
Antonis Karakottas\thanks{ankarako@iti.gr}\footnotemark[1],
Alexandros Papachristou\thanks{papachra@iti.gr}\footnotemark[1], 
Nikolaos Zioulis\thanks{nzioulis@iti.gr}\footnotemark[1], \\
Alexandros Doumanoglou\thanks{aldoum@iti.gr}, 
Dimitrios Zarpalas\thanks{zarpalas@iti.gr} and
Petros Daras\thanks{daras@iti.gr}}
\IEEEauthorblockA{Information Technologies Institute, Centre for Research and Technology - Hellas}
}

\def\fig{Figure }

\maketitle

\begin{abstract}
Multi-view capture systems are complex systems to engineer. They require technical knowledge to install and complex processes to setup. However, with the ongoing developments in new production methods, we are now at a position to be able to generate high quality realistic 3D assets. Nonetheless, the capturing systems developed with these methods are intertwined with them, relying on custom solutions and seldom - if not at all - publicly available. We design, develop and publicly offer a multi-view capture system based on the latest RGB-D sensor technology. We also develop a portable and easy-to-use external calibration process to allow for its widespread use.
\end{abstract}


\IEEEpeerreviewmaketitle

\section{Introduction}
	\label{sec:intro}
    
    The ongoing developments related to Virtual Reality (VR) and Augmented Reality (AR) technologies, and more importantly the availability of new presentation devices - head mounted displays (HMDs) - are also increasing the demand for new types of immersive media. Superseding traditional video, three-dimensional (3D) media are suited for both VR and AR and have been termed as "Free Viewpoint Video (FVV)" \cite{Collet2015}, "Volumetric Video", "Holograms" \cite{orts2016holoportation} and/or "3D/4D media"\footnote{These terms are used interchangeably in this document.}. They offer the ability of selecting any viewpoint to watch the content, essentially allowing for unrestricted viewing, therefore greatly increasing the feeling of immersion.
    
    Besides the expensive and laborious production of 3D media by artists using 3D modeling and animation software, there are various ways to 3D capture the real world and digitize it. Like typical video, 3D media can be consumed either in a live \cite{orts2016holoportation,alexiadis2017} or in an on-demand manner \cite{Collet2015,Robertini2017}, with state-of-the-art systems allowing for deformations and topology changes. Offline systems typically use a pre-defined template that will be fit on the data \cite{Theobalt2013} or otherwise utilize lengthy reconstruction processes \cite{Collet2015}. Consequently, 3D media production can either be real-time or post-processed. Either way, the backbone of realistic 3D content productions is a multi-view capture system. Such systems are complex to develop due to the large number of choices associated to their design. This system complexity is also translated to increased costs, specialized hardware (HW) requirements and technically demanding processes.
    
    Initially the multi-view capture system needs to be set up, a process that, depending on choice of the type and number of cameras/sensors can greatly vary. Using stereo pairs for the extraction of depth in a live setting requires extra processing power to be allocated for the disparity estimation task for each viewpoint (i.e. stereo pair) \cite{orts2016holoportation}. An offline system that operates on a template model fitting basis using the extracted silhouettes \cite{de2008performance, gall2009motion}, requires a larger number of cameras whose live feeds need to be recorded, thereby necessitating the use of large storage. The most suitable topology and architecture depends on the targeted use case. In the former case, besides setting up the stereo pairs, each one needs to be connected to a computer, with the processing offloaded to another workstation. In the latter case, depending on the frame-rate, resolution, encoding performance and disk writing throughput, the setup of a multi-disk server or a distributed local storage topology is required. 
    
    Following the installation of the multi-view capture system, a number of preparatory steps are needed before its actual use. These potentially involve spatial (external and internal calibration) and temporal (synchronization) alignment of the sensors. These processes can introduce new HW requirements (e.g. external signal triggers for synchronization \cite{de2008performance,orts2016holoportation,Collet2015}, or external optical tracking systems for calibration \cite{beck2017sweeping}) and are usually accomplished via complex procedures (e.g. moving checkerboard \cite{orts2016holoportation} or intricate registration structures \cite{Collet2015}). 
    
    Overall, as a combination of design decisions and complexity in operating, most existing multi-view capture systems are hindered by high HW costs, stationarity due to being hard to relocate after installation, or come with considerable technical requirements, forbidding adaptability and non-expert use. Our goal in this work is to design and deliver a flexible and up-to-date consumer level multi-view capture system to support affordable content creation for AR and VR. Our design is oriented towards taking steps in improving cost expenditure, portability, re-use and ease-of-use. In summary, our contributions are the following:
    
    \begin{itemize}
    	\item{A publicly available volumetric capture system utilizing recent sensor technology offered online at \url{https://github.com/VCL3D/VolumetricCapture}.}
        \item{The design of a low-cost, portable and flexible multi-view capture system.}
        \item{A quick, robust, user friendly and affordable multi-sensor calibration process.}
    \end{itemize}
\section{Related Work}

Multi-view capturing systems have mostly been developed for eventually producing three-dimensional content and are highly complex systems to design \cite{kubota2007multiview}. They typically require numerous sensors that need to be positioned, synchronized and calibrated and functionally they need to support either, or both, live acquisition and recording. They capture full 3D, by extracting the geometrical information of the captured scene, or pseudo-3D, by estimating the scene's depth and offering limited free viewpoint selection. Two of the pioneering works in this direction are \cite{kanade1997virtualized} and \cite{kauff2002immersive} respectively. The first one used a large number of cameras placed in a dome to surround the captured area and extracted complete geometric information, while the second one placed the cameras in front of the users and estimated the captured scene's depth. 

A state-of-the-art multi-view capturing dome has recently been presented in \cite{joo2017panoptic} that comprises 480 VGA, 31 HD cameras and 10 Microsoft Kinect 2.0. Its primary design goal was the social capture of multiple people. The system is calibrated using structure from motion and bundle adjustment using a white tent with a pattern projected on it. While being a very impressive system to engineer, it is nonetheless a very rigid, complex and expensive one. A more recent work for frontal facing multi-view capture \cite{lou2005real} showcased 32 cameras placed in an arc configuration which was calibrated by matching features found on the floor without the use of any pattern. Similarly, a system of 18 cameras in an array configuration that also leveraged the power of GPUs for real-time 3D reconstruction was presented in \cite{marton2011real}. However, its calibration was accomplished by using Tsai's checkerboard method \cite{tsai1987versatile}, a complex and cumbersome process which requires technical knowledge by the operator.

For full 3D capture, model-based performance capture methods  \cite{ de2008performance, Robertini2017, gall2009motion} allowed for the reduction of the number of sensors, compared to the aforementioned dome placement approaches, by employing 8 cameras perimetrically pointing inwards. As depth sensors' quality started improving, their deployment in multi-view systems quickly followed as a way to address the issues of camera-based capturing systems, namely low 3D reconstruction quality and green screen requirements. However, preliminary attempts were still calibrated using the inefficient checkerboard process \cite{kim2008design}, limiting their flexibility.

As commercial grade depth sensors, and more importantly, integrated color and depth (RGB-D) sensors started becoming available, a surge of renewed interest in 3D real-time or 4D post-processed content production quickly followed. Nonetheless, preliminary systems using multiple Kinect sensors either for 3D reconstruction \cite{ahmed2014using} or marker-less motion capture \cite{berger2011markerless} still used checkerboard based calibration approaches with even custom materials required for the latter one. However, in \cite{berger2011markerless}, an initial attempt in taking a step ahead of the typical calibration process was made by offering an alternative calibration process using a moving point light source. At the same time, structure-based calibration systems started surfacing typically using markers to either directly estimate each camera's pose with respect to the structure \cite{omnikinect2012,Zioulis2016} or as initial estimates to be densely refined \cite{kowalski2015live}. However, even the state-of-the-art real-time 3D capturing system of \cite{orts2016holoportation}, using 16 near infrared cameras, 8 color cameras and 8 structured light projectors, still relies on the checkerboard method of \cite{zhang00} for calibration. Similarly, the high quality 4D production system of \cite{Collet2015} that consists of  106 cameras, relies on an octagonal tower structure for its calibration, albeit being very complex and hard to assembly and re-locate.

As a result there have been various works aiming to make the overall calibration process easier. The work of \cite{beck2017sweeping} utilized an expensive external optical tracking system to calibrate the multi-view system's captured volume area using a checkerboard to further improve the accuracy of the solution and achieve an easier and more robust work flow. In \cite{fornaser2017automatic} and \cite{su2018fast}, the authors utilize a colored ball that is moved within the capturing area to establish correspondences and calibrate the multi sensor systems. Additionally, in \cite{fornaser2017automatic}, their method simultaneously synchronizes the sensors in addition to calibrating them. While HW synchronization is the optimal, some sensors do not support it, necessitating the use of software (SW) based synchronization approaches. More recently, \cite{wscg2018} presented a marker-less structure-based multi-sensor calibration using a CNN trained with synthetic structure renders. However, training was limited to specific angle intervals around the structure. Nonetheless, the presented multi-sensor calibration process was made significantly easier. 

Overall, we find that most systems require complex processes to calibrate that need heavy human operations - usually with technical knowledge. This renders them hard to (re-)use for commercial purposes, due to heavy customization in materials and configurations, also limiting their portability. In addition, most - if not all - systems' implementations are not publicly available with some being notorious hard to assemble and/or develop. Our goal is to design and develop, and publicly offer an easy to setup multi-view capturing system, with low-cost components and minimize the technical requirements as well as process complexity in operating it.

\section{Volumetric Capture}
	\label{sec:volCapArch}
    
\begin{figure*}[!t]
	\centering
    \subfloat[]
    {
    	\includegraphics[scale = 0.24]{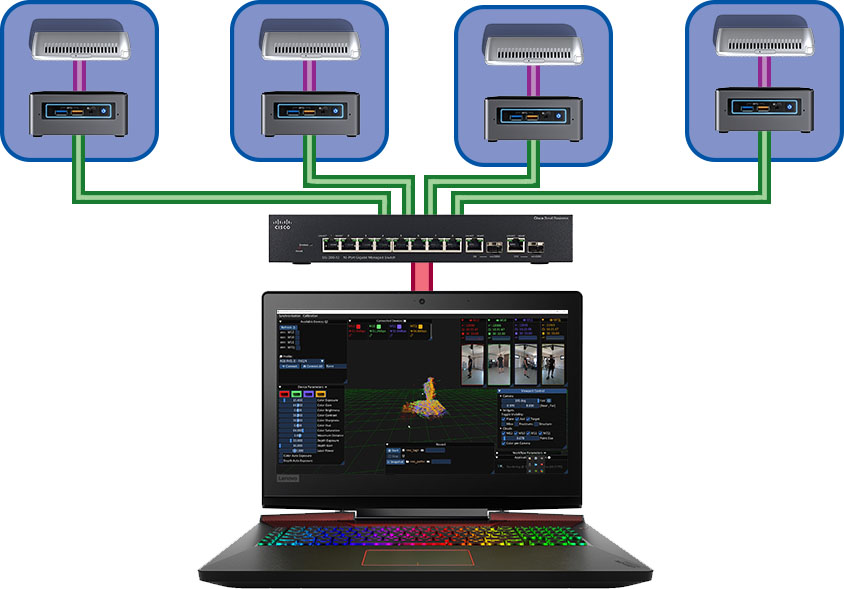}
        \label{subfig:overview}
    }
    \hspace{\fill}
    \subfloat[]
    {
    	\includegraphics[scale = 0.28]{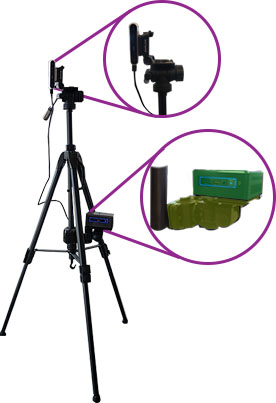}
        \label{subfig:acq}
    }
    \hspace{\fill}
    \subfloat[]
    {
    	\includegraphics[scale = 0.7]{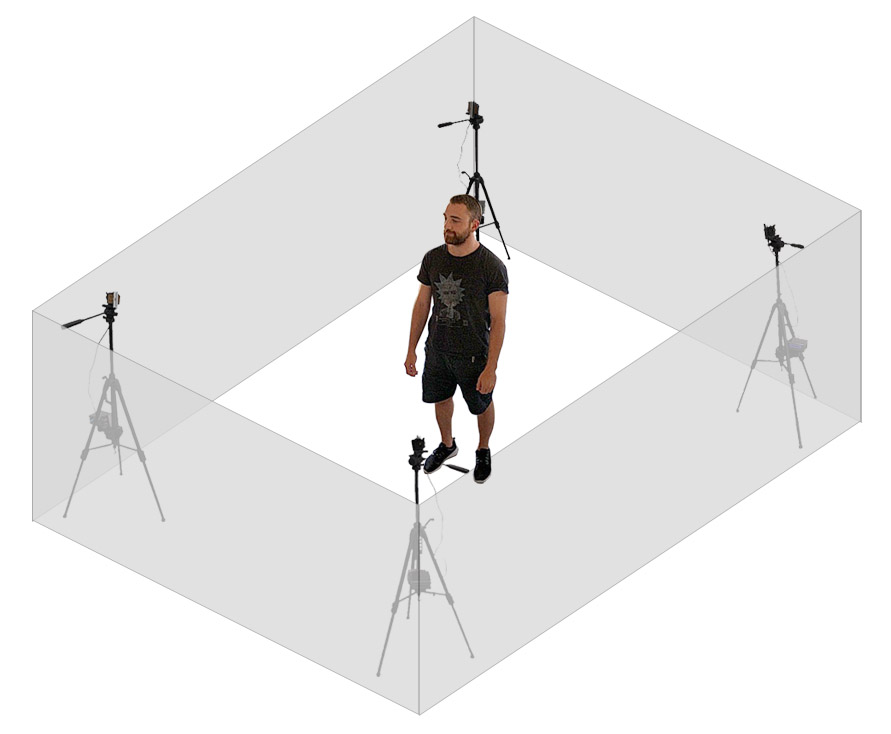}
        \label{subfig:setup}
    }
    \label{fig:hardArch}
    \caption{Capturing System Overview and Architecture. (a) Our basic system setup, utilizes $N=4$ acquisition modules (\textit{eyes}) and a central \textit{orchestrator} workstation. The \textit{orchestrator} communicates with the \textit{eyes} through LAN. (b) The acquisition module is composed of an Intel RealSense D415 sensor mounted on a tripod, connected to an Intel NUC processing unit, also mounted on the same tripod. (c) Example volumetric capturing station setup with the sensors looking inwards and capturing a $360^o$ view of the subject.}
\end{figure*}

    Our volumetric capture system is designed to orchestrate the capturing, streaming and recording of the data acquired from a multi-sensor infrastructure. While in principle it can be used for moving sensors too, our focus is oriented towards static inwards placement for capturing human performances within a predefined space. Our design choices strive to reach an optimal balance among affordability, modularity, portability, scalability and usability.
    
    \textbf{Sensor:} We employ the most recent version of the Intel RealSense technology \cite{keselmanintel}, a consumer-grade RGB-D sensor which allows us to reap the advantages of integrated depth sensing. This reduces the complexity of our system as we can deploy a single integrated RGB-D sensor instead of 4 (2 gray-scale for stereo computation, 1 for color acquisition and 1 projector to improve depth estimation in uniform colored regions) as in \cite{orts2016holoportation}. In addition, compared to approaches surrounding the captured area with monocular sensors \cite{Theobalt2013,gall2009motion,Robertini2017}, we can deploy less number of sensors due to the availability of depth information. More specifically we use the D415 sensor\footnote{\url{https://software.intel.com/en-us/realsense/d400}}, which compared to its sibling, the D435, offers better quality at closer distances due to a denser projection pattern and also supports HW synchronization between its color and depth sensors. In addition, this type of sensors offer inter-sensor HW synchronization. Contrary, using Microsoft Kinects, would require a soft synchronization solution, that are typically SW-based, like the audio synchronization of \cite{alexiadis2017}, adding yet another process when setting up the system. Further, the D415 sensors allow for setting up each sensor as a master or slave, and as a result, the requirement and added complexity and cost of using and having to setup, external HW triggers is lifted.
    
\begin{figure}[!h]
	\centering
    \includegraphics[scale = 0.43]{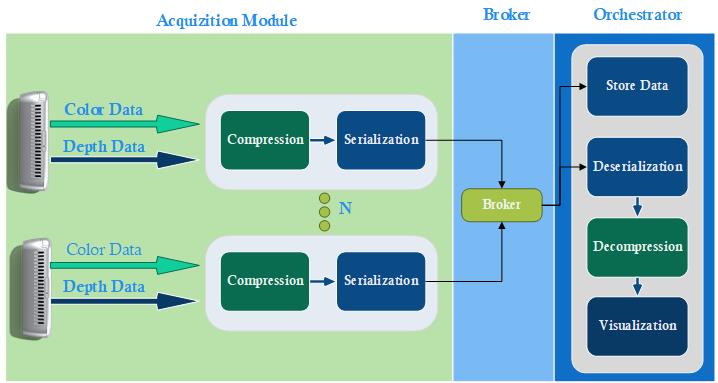}    
    \caption{\textit{Volumetric Capture} data flow. Multiple ($N$) acquisition modules (\textit{eyes}) capture the scene's color and depth information. The acquired data are first compressed, serialized and published to the message \textit{broker} over the network. The \textit{orchestrator} client then deserializes and decompresses the received messages to visualize and/or store them.}
    \label{fig:dataFlow}
\end{figure}

    \textbf{Architecture:} Our building block is an acquisition module, called an \textit{eye}, that represents a viewpoint positioned globally in relation to the capturing volume and is serving a RGB-D data stream. We connect $N$ \textit{eyes} in a distributed fashion to work towards a common goal, providing fused colored point clouds or otherwise registered multi-view RGB-D streams. These are delivered into a client that is also the \textit{orchestrator}, controlling the behavior and parameterization of the \textit{eye} server units through message passing. Control messages as well as data streams are transferred by a \textit{broker} using a publish-subscribe event-based architecture, with the system's data flow depicted in Figure \ref{fig:dataFlow}. All these aforementioned components comprise a single coherent, \textit{Volumetric Capture} system.
    
    \textbf{Hardware:} The physical interpretation of our \textit{eye} acquisition module is illustrated in Figure \ref{subfig:acq}. A D415 sensor is mounted on a tripod and connected to an Intel NUC mini-PC, which is in turn mounted on a tripod VESA mount. These, and the \textit{orchestrator}, are connected via Ethernet cables to a LAN switch as seen in Figure \ref{subfig:overview}. The switch's bandwidth depends on the number of sensors and their streams' resolution and frame rates, but for typical $360^o$ capture use, at least a $1$ Gbps bandwidth is required. Another important specification is that it needs to be non-blocking to be able to handle all of its ports' bandwidths  at full capacity simultaneously. This is essential when using HW synchronization, as network traffic comes in bursts that would otherwise manifest in extra latency. Furthermore, through the use of mini-PCs, we distribute processing at a negligible effect on the system's portability. This way we move the computational burden of compression and pre-processing on the acquisition modules, allowing for more efficient recording and overall reduced computational complexity on the receiving client.
       
   An alternative to our distributed design would be to connect all sensors into a single workstation, which could arguably slightly increase its portability. However, this design choice requires the installation of additional USB 3.0 controllers, as each sensor consumes high bandwidth to stream data in higher resolutions and frame rates. Because of this, the cables of the D415s are very short ($1m$) and, therefore, high quality USB 3.0 extension cables would be required. Depending on the distance and the data rate, optical repeaters might be needed that greatly increase the cost, bringing it on par with our HW choices. Further, scalability would be limited to the USB 3.0 extension slots that a high-end motherboard can support and input-output bandwidth. 

   \textbf{Implementation Details: } Our system's main components, the client (\textit{orchestrator}) and server (\textit{eye}), are natively implemented in C++. Since we utilize headless clients (mini-PCs), an automated way of discovering the acquisition modules is required. To that end, we deploy a service to each mini-PC, developed in C\#, that listens for broadcast messages to spawn the \textit{eye} component process. For our message broker, we use RabbitMQ which can be co-located with the \textit{orchestrator} component. We use lossy compression for the color streams and lossless compression for the depth streams. Compression method choices aim at minimizing acquisition latency to enable use in real-time 3D reconstruction scenarios. To that end, we use intra-frame JPEG compression for the color streams, and entropy-based compression for the depth streams. For the former an SIMD optimized version \cite{turbojpeg} is used, while for the latter, a variety of algorithms are used under a blocking optimization technique \cite{blosc}. This allows for a more explicit control of the overall bandwidth that each \textit{eye} unit produces, as the depth stream mostly dominates the encoding performance and resulting compressed frame sizes.    
   
\begin{figure}[!h]
	\centering
    \includegraphics[width = \columnwidth]{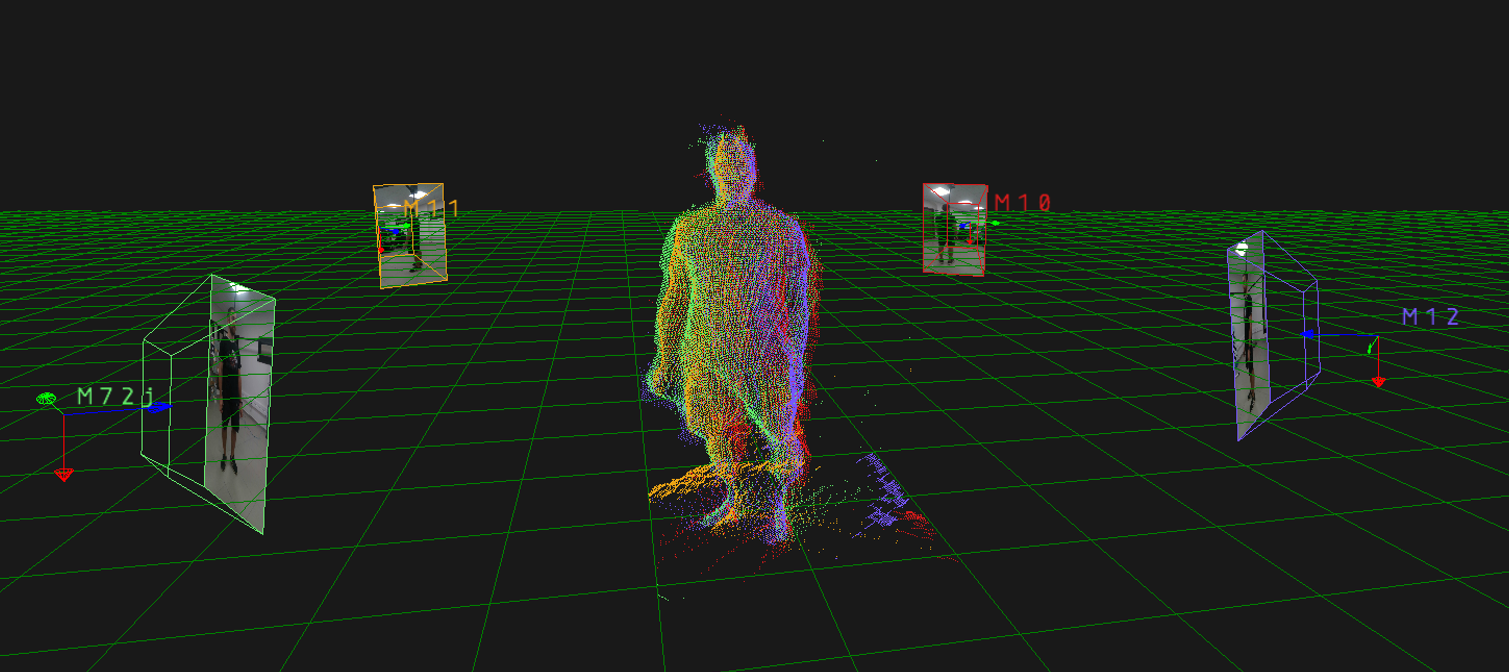}
    \label{fig:acqMod}
    \caption{3D capture snapshot acquired from the \textit{Volumetric Capture} application showcasing the calibrated output when capturing a human subject. Each viewpoint's pose is also depicted via the camera frustum placements.}
\end{figure}

\section{Practical Calibration}

\begin{figure}[!ht]
	\centering
    \includegraphics[scale = 0.18]{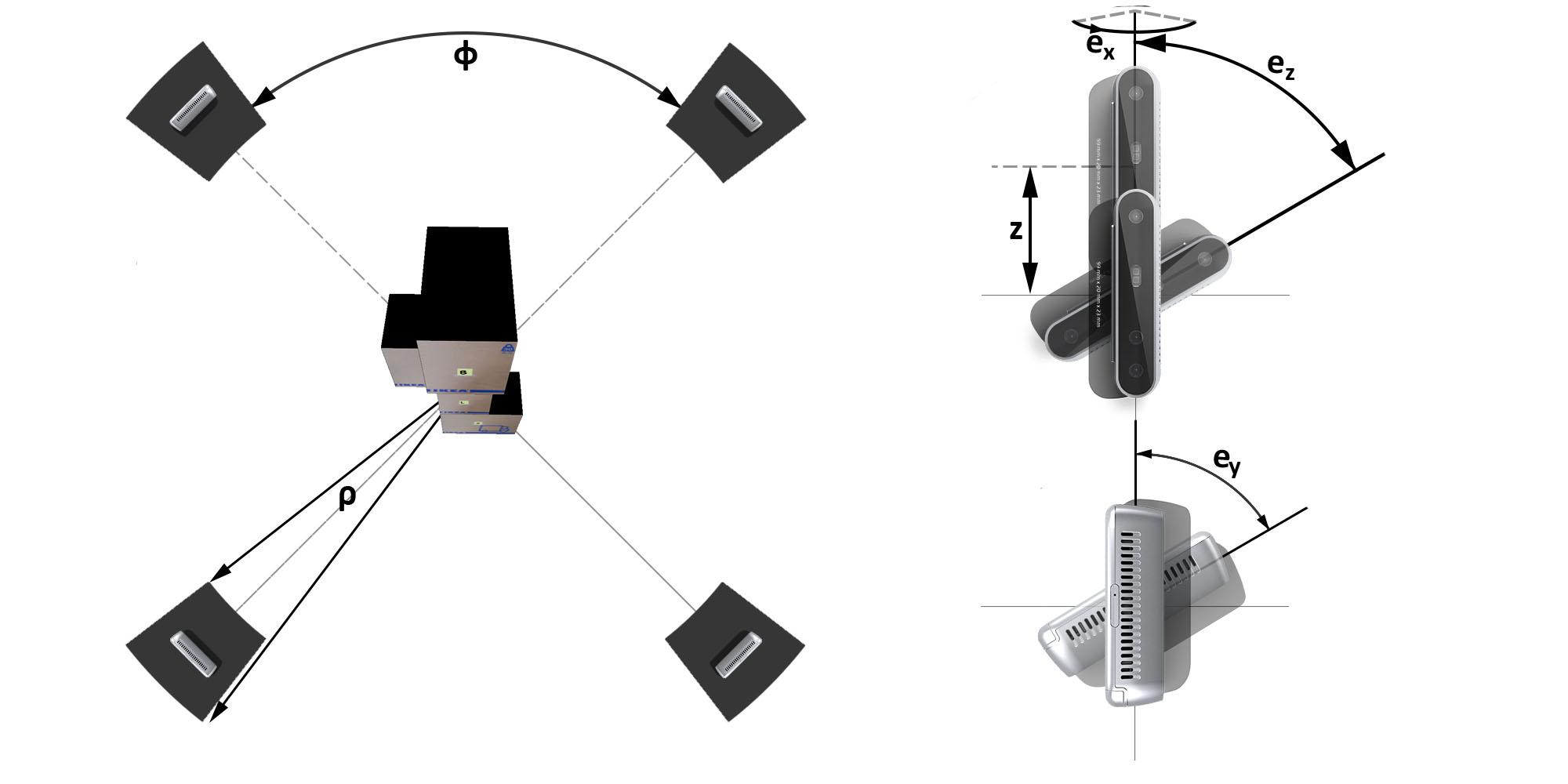}
    \caption{Pose generation sampling parameters illustration. Each sample pose is randomly generated at the $(\rho, \phi, z)$ cylindrical coordinate defined around the calibration structure's origin. It is further randomly rotated around the $(\mathbf{x}, \mathbf{y}, \mathbf{z})$ axii at a respective Euler angle  $(e_{\mathbf{x}}, e_{\mathbf{y}}, e_{\mathbf{z}})$.}
     \label{fig:trainStruct}
\end{figure}

The cornerstone of multi-view systems is the spatial alignment, or otherwise external calibration of the sensors with respect to a global coordinate system, as seen in Figure \ref{fig:trainStruct}. Typical checkerboard calibration processes require heavy human intervention as well as technical knowledge to avoid ambiguous or error-prone checkerboard poses. In order to make this process more convenient and usable by non-technical personnel, we opt for a structure-based calibration that only requires users to assemble and place the structure within the capturing volume. While previous such approaches placed markers or patterns on the structures \cite{livescan3d,alexiadis2017,Collet2015}, we extend and improve the marker-less calibration of \cite{wscg2018}.

\begin{figure}[!h]
	\centering
    \subfloat[]
    {
    	\includegraphics[scale = 0.43]{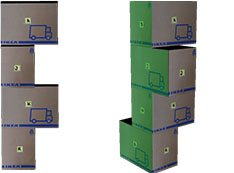}
        \label{subfig:oldstruct}
    }
    \subfloat[]
    {
		\includegraphics[scale = 0.43]{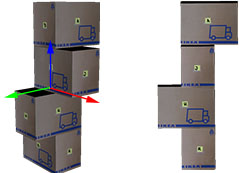}    
        \label{subfig:newstruct}
    }
    \subfloat[]
    {
    	\includegraphics[scale = 0.095]{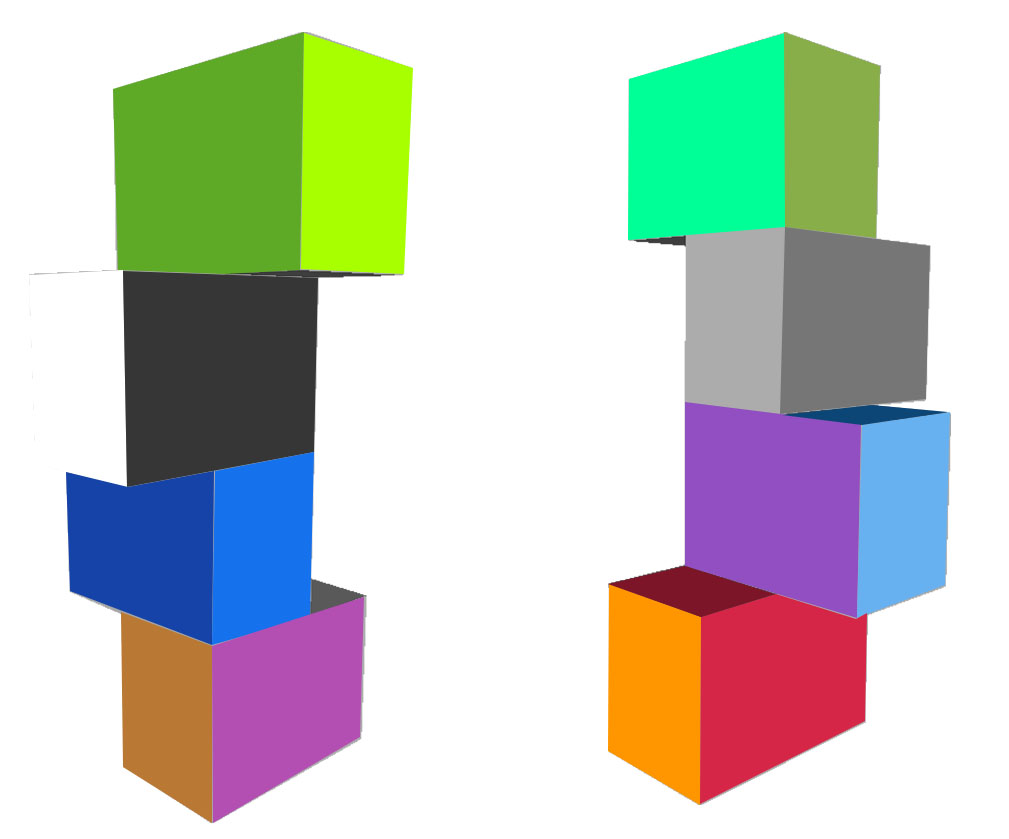}
        \label{subfig:segments}
    }
    \caption{Update of the calibration structure. In \protect \subref{subfig:oldstruct} the old calibration structure is presented, on which the planar side can be sheen with green overlay. In \protect \subref{subfig:newstruct} where the updated calibration structure is presented, there is no longer any coplanar side. Each side segments of the calibration structure can be seen in \protect \subref{subfig:segments}}
        \label{fig:boxchanges}
\end{figure}

\textbf{Structure: }Similar to \cite{alexiadis2017} and \cite{wscg2018} we use a structure assembled out of commercially available packaging boxes whose dimensions are standardized. This allows us to create a virtual replica of the calibration structure in the form of a 3D model. In practice we use 4 boxes and deviate from the structure assembly of previous approaches so as to create a fully asymmetric structure that, at the same time, has no fully planar views. This way, we naturally resolve any difficulties in identifying each of the structure's sides and further guarantee that the extracted correspondences will not produce ill-formed or ambiguous solutions when used to estimate the sensor's pose. The updated structure can be seen in Figure \ref{fig:boxchanges} that also showcases the changes compared to the structure of \cite{wscg2018}.

\textbf{Training Data: }Our goal is to use the structure's prior knowledge to establish correspondences between each sensor's viewpoint and the global coordinate system that the structure defines. Since we aim to be using no markers, and therefore no color information, this is accomplished by training a CNN to identify these correspondences. The virtual 3D model can then be used to generate training pairs on-the-fly. By placing a virtual camera at a relative position around the 3D model that defines the center of the coordinate space, we can render it and generate a depth map $D(\mathbf{p}) \in \mathbb{R}$ out of the resulting $z$-buffer, where $\mathbf{p}=(u, v) \in \Omega \subset \mathbb{N}^2$ represents pixel coordinates in the image domain $\Omega : u \in [1, \dots, W], v \in [1, \dots, H]$, with $W$ and $H$ its width and height respectively. Given also a material of the model, we can additionally output a texture map $\mathbf{L}(\mathbf{p}) \in \mathbb{R}^3$ acquired from the resulting render buffer. By assigning a different material (i.e. color) in each of the four boxes' sides (total $24$ distinct sides), these images then correspond to a depth map semantic segmentation supervision pair $\{D(\mathbf{p}), \mathbf{L}(\mathbf{p})\}$. Our rendered data are generated at a resolution of $320\times180$, corresponding to a downscaled (factor of $4$) depth map of a D415 sensor. We also add noise to the resulting depth maps and augment them with random backgrounds as in \cite{wscg2018}, later denoted as $\tilde{D}(\mathbf{p})$. However, our approach differs in various ways that will be thereafter explained.

\textbf{Pose Sampling: }We sample poses using cylindrical coordinates $\mathbf{t}_c = (\rho, \phi, z)$ defined on the virtual structure's coordinate system. These are then transformed to sensor poses as follows: i) we extract a Cartesian 3D position $\mathbf{t}_{3D}$ from each $\mathbf{t}_c$; ii) we estimate a rotation matrix $\mathbf{R}$ by estimating the view matrix from $\mathbf{t}$ to the origin $(0,0,0)$ of the coordinate system - which is at the center of the virtual structure model - using the $\mathbf{y}$ axis as the up vector; iii) we augment the rotation $\mathbf{R}$ by adding rotational noise via the composition of random rotations $\mathbf{R}_{\mathbf{i}}, \mathbf{i} \in \{\mathbf{x}, \mathbf{y}, \mathbf{z}\}$ around each axis. Similar to \cite{wscg2018}, we sample these variables from uniform distributions $U(a,b,c)$ in intervals $[a,b]$ at steps $c$:
\begin{align}
	\centering
	\label{eqn:uniform}
	\begin{split}
    \phi_n \sim \mathit{U}(n\times\frac{360^o}{N}-10^o, n\times\frac{360^o}{N}+10^o, 2.5^o),
	\\
 	z \sim \mathit{U}(0.28m, 0.7m, 0.02m),
	\\
 	\rho \sim \mathit{U}(1.75m, 2.25m, 0.02m),
	\\
 	e_{\{\mathbf{x}, \mathbf{y}, \mathbf{z} \}} \sim \mathit{U}(-10^\circ, 10^\circ, 2.5^\circ),    	
\end{split}
\end{align}
with $n \in [1,\dots,N]$ and $e$ being a Euclidean angle around axii $\{\mathbf{x}, \mathbf{y}, \mathbf{z}\}$ that is transformed into a rotation matrix $\mathbf{R}_{\{\mathbf{x}, \mathbf{y}, \mathbf{z}\}}$. An illustration of this sampling is available in Figure \ref{fig:trainStruct}. Using this sampling we try to cover for a wide range of placements of each $n_{th}$ sensor in a variety of capturing scenarios of $N$ sensors, while also modeling realistic imperfect approximate positioning. Contrary to \cite{wscg2018}, we sample across the whole circle around the structure but enforce that groups of $N$ sensors will be placed approximately at the appropriate $\phi$ angle intervals.    

\textbf{Network: }Instead of training a CNN to predict dense labels to identify each specific box's side on a per depth map basis, we exploit the complementarity of the $N$ viewpoints and train our CNN to receive as input all viewpoints jointly. As the goal is to achieve $360^o$ coverage around a capturing volume, each viewpoint's depth view is related to the other viewpoints. Given that the viewpoints will be evenly placed around the structure, each viewpoint's depth map is complementary input to the rest as it will restrict their predictions. Consequently, we design a CNN that receives $N$ depth map inputs $\tilde{D}(\mathbf{p})$ and fuses their information to extract this relative and complementary information. As we cannot fix the sequence of the inputs because it requires knowledge of the spatial relations, which is our final objective, we randomize the order of the inputs the CNN receives during training.

\begin{figure*}[!ht]
	\centering
    \includegraphics[width = 0.9\textwidth]{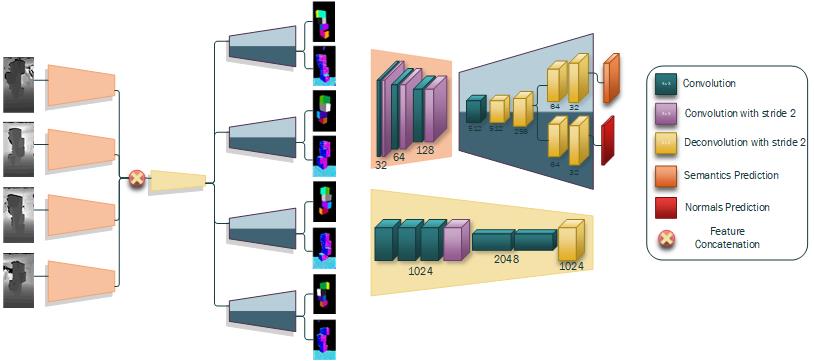}
    \caption{Our CNN architecture comprises 4 input encoding branches, one for each view, a bottleneck, and 4 output decoding branches. The input branches' features are fused through concatenation and fed into the bottleneck. The four output branches decode the bottleneck into two separate predictions for each branch, densely estimating a normal and label map for each branch. Each input branch comprises three blocks having two convolution (\textit{conv}) layers each, with the second downscaling its output features (stride equal to 2). The bottleneck contains three blocks. The first comprises four \textit{convs}, with the last one downscaling its output, while the second block comprises two \textit{convs}. The bottleneck's third block updates its input features by using a deconvolution (\textit{deconv}) layer (stride equal to 2). Each output branch contains four blocks of layers. The first block utilizes one \textit{conv}, followed by two \textit{deconvs}. The last \textit{deconv's} output branches out, to feed the two internal branches of the output branch. Both of these are composed by two \textit{deconvs}, with the only difference being the number of the predicted features. The segmentation branch's prediction layer classifies 25 features while the normal prediction branch produces 3 output features. All (de-)convolutional layers use $3\times3$ kernels.}
     \label{fig:structNet}
\end{figure*}

\textbf{Multi-task Learning: }Our task is to label each one of the structure's box's sides, all of which are planar surfaces. Taking into account that the planar surfaces' orientation is defined by their normal, it is apparent that the observed scene's normal information is complementary to our box side labeling task. We exploit this complementarity by designing our network for multi-task learning to take advantage of this inter-task relationship. During each render, we also output a normal map $\mathbf{N}(\mathbf{p}) \in \mathbb{R}^3$ in another render-buffer that will be used to supervise the CNN's secondary task, normal estimation from depth maps. Summarizing, our CNN, with its architecture presented in Figure \ref{fig:structNet}, receives as input multiple - randomly perturbed - depth maps $\tilde{D}_i(\mathbf{p}), i \in \ [1, \dots, N]$ observing the same structure from different viewpoints, as well as their ground truth label maps $L_i$ and jointly estimates the semantic labels' probability distributions $\hat{\mathbf{L}}_i(\mathbf{p})$ of the structure's boxes' sides as well as their normal maps $\hat{\mathbf{N}}_i(\mathbf{p})$, for each input depth map, while fusing their multi-view information. We use a cross-entropy loss for the labels and a $\mathit{L}_2$ loss for the normals. 

We try to minimize the overall loss:
\begin{equation} 
\mathit{E_{overall} = \sum_{i=1}^{N} \mathit{E_i}}, \, \mathit{E_i} = \mathit{E_{normal}} + \lambda \mathit{E_{semantic}}
\end{equation}
over all dataset samples with:
\begin{equation}
\mathit{E_{normal}} = \dfrac{1}{M} \sum_{\mathbf{p}}^{\Omega} ||\hat{\mathbf{N}}(\mathbf{p}) - \mathbf{N}(\mathbf{p}) ||^2,
\end{equation}
\begin{equation}
\mathit{E_{semantic}} = \dfrac{1}{M} \sum_{\mathbf{p}}^{\Omega} \mathit{Pr}(L(\mathbf{p})) \log(smax(\hat{\mathbf{L}}(\mathbf{p}))),
\end{equation}
where $\lambda$ is a weight factor balancing the contributions of the regression $\mathit{E_{normal}}$ and classification $\mathit{E_{semantic}}$ losses, $M = W \times H$ equals the total number of pixels, $\mathit{Pr}$ is a function that extracts the ground truth probability distribution from the rendered texture map $\mathbf{L}$, for each pixel $\mathbf{p}$, and $smax$ is the softmax function evaluated at each corresponding pixel $\mathbf{p}$.

We do not enforce normalized predictions as it has been observed that the $\mathit{L}_2$ loss alone will suffice in producing normalized values \cite{kendall2015posenet}.

\textbf{Refinement: }Finally, we refine the dense label predictions of the CNN using a dense fully-connected Conditional Random Field (CRF) model \cite{krahenbuhl2011efficient} formulated over the predicted label distributions $\hat{\mathbf{L}}$ and normals maps $\hat{\mathbf{N}}$. The per pixel $\mathbf{p}$ energy function is:
\begin{equation}
E_{CRF}(\mathbf{p}) = \sum_{i} \psi_{unary}(\mathbf{p}_i) + \sum_i \sum_{j \neq i} \psi_{pairwise} (\mathbf{p}_i, \mathbf{p}_j).
\end{equation}
The \textit{unary} potential $\psi_{unary}(\mathbf{p})$ is the densely predicted distribution over the label space, describing the cost of a pixel taking the corresponding label as estimated by the CNN's output label probability distribution map $\hat{\mathbf{L}}$. The \textit{pairwise} potential terms are Gaussian edge potentials, describing the cost of variables $i$ and $j$ taking their corresponding labels respectively. For the pairwise term we use as a feature formulation $\mathbf{f}$ similar to \cite{krahenbuhl2011efficient} by including the positions in the image space, but instead of the values in the image (RGB) domain, we use the values of the predicted normal map $\hat{\mathbf{N}}$. Therefore the appearance kernel of \cite{krahenbuhl2011efficient} becomes:
\begin{equation}
\alpha(\mathbf{f}_i, \mathbf{f}_j) = exp(-\frac{| \mathbf{p}_i - \mathbf{p}_j |^2}{2\sigma_{2D}^2} - \frac{| \hat{\mathbf{N}}(\mathbf{p}_i) - \hat{\mathbf{N}}(\mathbf{p}_j) |^2}{2\sigma_{3D}^2}),
\end{equation}
where $\sigma_{2D}$ and $\sigma_{3D}$ the ranges that the spatial and normal kernels operate respectively. This is based on a the same intuition, that since we are labeling planar surfaces, the estimated normals define the labels, and therefore, their edges and similarities help in improving the resulting label predictions.

\textbf{Correspondences and Optimization: } Once we obtain our refined labels, we can extract a single correspondence from each labeled region. Similar to \cite{wscg2018}, we back-project the depths for each region and obtain 3D coordinates, from which we extract their median value to obtain a robust estimate for each box's side's mid-point. We can then obtain an initial estimate of each sensor viewpoint's pose via Procrustes analysis \cite{kendall1989survey} using the 3D-to-3D correspondences between the sensor's view and the virtual 3D model. Using this initial estimate, we then optimize the dense point clouds from each back-projected depth map, using ICP formulated with a point-to-plane error, under a graph-based optimization in order to obtain a global solution (more details can be found at \cite{wscg2018}).

\section{Results}

We evaluate our proposed calibration method under a variety of different sensor placements to showcase its  robustness to user sensor placement. Our system can support an arbitrary number of sensors, limited by the HW (network speed, HDD write speed) and the use case requirements (resolution, frame rate). However, in our evaluation we focus on a $N=4$ sensor setup that achieves optimal coverage while keeping the HW requirements - and by extension the cost - to a minimum. We train our network using Caffe \cite{jia2014caffe} on an Nvidia Titan X, with $N=4$ input depth maps, for $150k$ iterations and an initial learning rate of $10^{-3}$, using the ADAM optimizer \cite{kingma2014adam} initialized with its default values, and with $\lambda = 0.1$. When rendering we use average depth sensor intrinsic parameters obtained by 9 different factory D415 sensors, divided by the downscaling factor. We test our network's performance by generating a test set with pose samples drawn from uniform distributions with different parameters than those reported in Equation \ref{eqn:uniform}. These are selected to produce labeled images from different pose configurations than those used for training. In this test set, our model achieves a $96.17\%$ mIoU.

We compare our calibration results against other similar methods, both structure-based and object-based. For the structure-based method comparisons we use LiveScan3D \cite{kowalski2015live} by attaching their markers on our structure, and we additionally compare against the  marker-based method of \cite{alexiadis2017,Zioulis2016} by also attaching QR codes on our structure and enhancing it with the same graph-based dense optimization step we use, effectively evaluating only the correspondence extraction's effect for the initial pose estimation. For object-based method comparisons we use the approaches of \cite{fornaser2017automatic} and \cite{su2018fast} which utilize a ball that is moved within the capturing volume to establish correspondences to then optimize the sensors' poses. In order to use the same sequences for comparison, we updated the method of \cite{su2018fast} to work with a green colored ball. Therefore, we first capture RGB-D data by moving a green ball attached on a stick with a known diameter ($20cm$) within the capturing volume, and then we place the structure and re-capture data to obtain the necessary input for all methods. We conduct these experiments for 5 different placements as presented in Table \ref{tab:semantic}. For evaluating the accuracy of the calibration methods we use the Rooted Mean Squared Euclidean (RMSE) distance between the closest points of adjacent views. The final error metric of each method is extracted by taking the mean RMSE distance of all pairs of adjacent views. 

\begin{table}[!h]
	\centering
    \caption{RMSE results (in mm) of our method and the compared ones. \textnormal{Approximate sensors' placements were: $a \sim \{\rho\text{ : }1.9m, z\text{ : }0.38m$\},
			$b \sim \{\rho\text{ : }1.3m, z\text{ : }0.38m$\},
			$c \sim \{\rho\text{ : }1.5m, z\text{ : }0.38m$\},
			$d \sim \{\rho\text{ : }1.5m, z\text{ : }0.38m-0.48$\}, and 
			$e \sim \{\rho\text{ : }1.8m, z\text{ : }0.38-0.48m$\}.
            For those not available, the methods did not manage to converge.).
	}}
	\label{tab:semantic}
	\begin{tabular}{c|ccccc}
		Method & a            & b            & c             & d             & e             \\ \cline{1-6} 
		\cite{kowalski2015live}       & 21.8          & N/A          & N/A          & N/A          & N/A	\\
       	\cite{fornaser2017automatic}       & 20.82          & 18.41          & 20.79          & 21.83          & N/A	\\
        \cite{su2018fast}       & 21.57          & N/A          & 18.52          & 20.67          & 21.54	\\
        \cite{alexiadis2017,Zioulis2016}    & \textbf{16.53}          & \textbf{14.65}          & \textbf{15.06}          & \textbf{15.45}          & N/A \\
		Ours & 17.57         & 15.41 & 17.26          & 16.85          & \textbf{19.83}
	\end{tabular}
\end{table}

From the results presented in Table \ref{tab:semantic}, we see that not all methods manage to consistently converge into a good solution apart from ours. In addition, while the marker-based approach produced better results in those placements that it managed to converge, it should be noted that it required a lot of parameter fine-tuning of the SIFT detection parameters, on a per-experiment basis, to extract matching features. In addition, our marker-less method produces comparable accuracy results. We can therefore conclude that our method robustly produces high quality external calibration results with minimal human intervention and technical knowledge.

\begin{figure}[!h]
	\centering
    \subfloat[]{
    	\includegraphics[scale = 1]{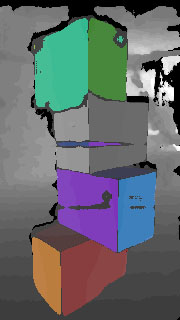}
        \includegraphics[scale = 1]{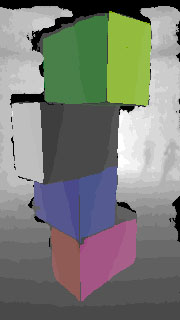}
        \includegraphics[scale = 1]{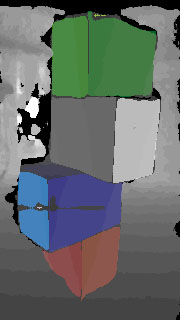}
        \includegraphics[scale = 1]{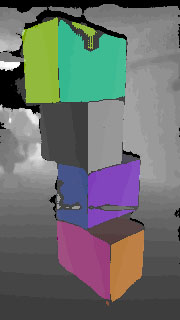}
        \label{subfig:pre}
    }
    \vspace*{-5pt}
    \subfloat[]{
    	\includegraphics[scale = 1]{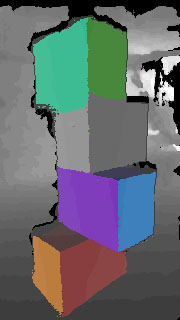}
        \includegraphics[scale = 1]{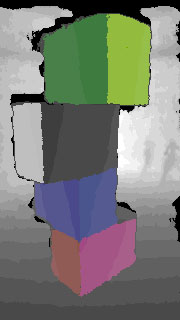}
        \includegraphics[scale = 1]{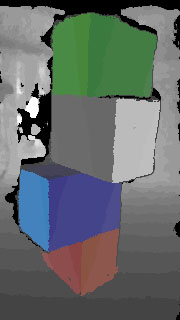}
        \includegraphics[scale = 1]{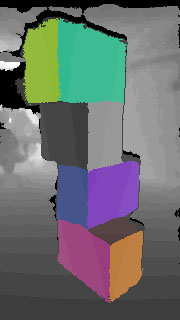}
        \label{subfig:post}
    }    
    \caption{Segmentation results before \protect \subref{subfig:pre} and after \protect \subref{subfig:post} applying the CRF post-refinement step.}
    \label{fig:prePostCRF}
\end{figure}

\begin{figure}[!h]
	\centering
    \subfloat[]{
    	\includegraphics[scale = 0.15]{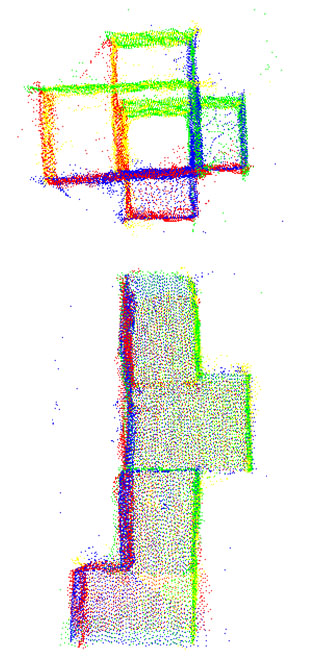}    
    }
    \subfloat[]{
    	\includegraphics[scale = 0.15]{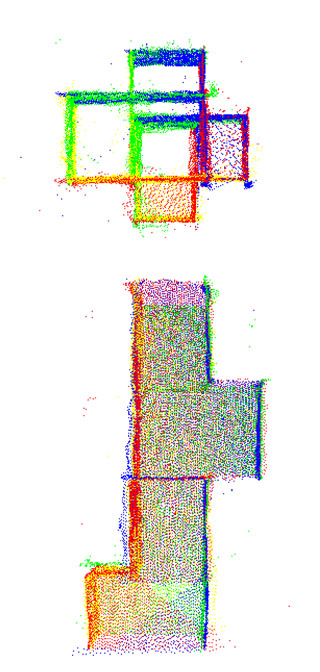}    
    }
    \subfloat[]{
    	\includegraphics[scale = 0.15]{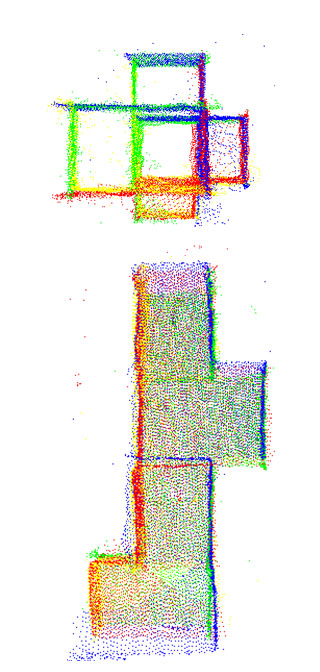}    
    }
    \subfloat[]{
    	\includegraphics[scale = 0.15]{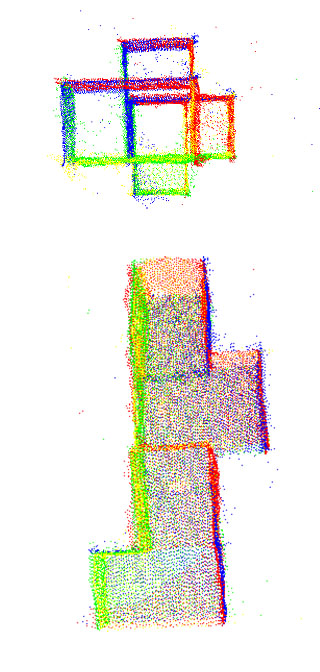}    
    }
    \subfloat[]{
    	\includegraphics[scale = 0.15]{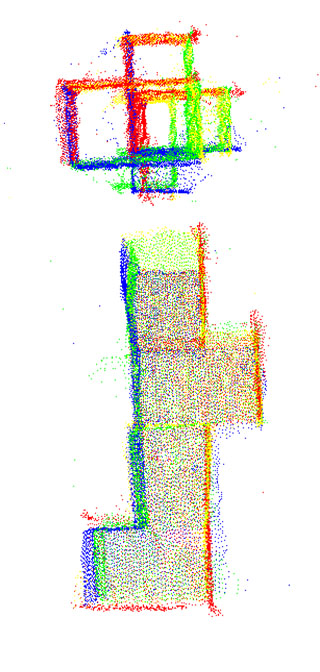}    
    }
    \caption{Qualitative results of the obtained external calibration among the four sensors' viewpoints in our five (a-e) experiments. Each viewpoint is colored with a different color. On the top row, we offer top-down views, while on the bottom row, their respective side views are illustrated.}
    \label{fig:exper}
\end{figure}

We additionally offer some qualitative results to showcase the effect of the post-refinement dense CRF step. Figure \ref{fig:prePostCRF} shows the output of the CNN for a quadruple of depth maps for experiment \textit{b}, and then presents the output of the post-refinement step that improves the quality of the labeled regions. This helps in establishing more accurate correspondences for the initial pose estimates and thus, better drives the subsequent dense graph-based optimization step. Moreover, we also  offer qualitative results of the accuracy of the registration for all the conducted experiments in Figure \ref{fig:exper}.

\section{Conclusion}
Multi-view capturing is gaining traction with the recent developments to AR and VR, however, multi-view systems are difficult to engineer and develop and are usually complex to setup and difficult to re-locate. We have designed and publicly offer a multi-view system based on recent sensor technology that is significantly lower cost, easier to setup and portable, in contrast with other systems in the literature. This was achieved through careful design decisions and the development of a new calibration method that is easy to use and at the same time, robust. Even though the demonstrated calibration process relies on learning a specific placement, this does not restrict for training new networks for other setups as well (e.g. 3 sensors at $120^o$ angle intervals, or 8 sensors arranged in two different 4 sensor perimeters at different heights). We believe that our system can be used as a basis future research on production methods, as well as for 3D content creation by freelancers and professionals alike enabling quicker workflows due to quicker and more flexible setup times. 
\section*{Acknowledgment}

This work was supported by the EU's H2020 Framework Programme funded project Hyper360 (GA~$761934$). We are also grateful for a hardware donation by Nvidia.

\bibliographystyle{IEEEtran}
\bibliography{bibliography}

\begin{thebibliography}{10}
\providecommand{\url}[1]{#1}
\csname url@samestyle\endcsname
\providecommand{\newblock}{\relax}
\providecommand{\bibinfo}[2]{#2}
\providecommand{\BIBentrySTDinterwordspacing}{\spaceskip=0pt\relax}
\providecommand{\BIBentryALTinterwordstretchfactor}{4}
\providecommand{\BIBentryALTinterwordspacing}{\spaceskip=\fontdimen2\font plus
\BIBentryALTinterwordstretchfactor\fontdimen3\font minus
  \fontdimen4\font\relax}
\providecommand{\BIBforeignlanguage}[2]{{%
\expandafter\ifx\csname l@#1\endcsname\relax
\typeout{** WARNING: IEEEtran.bst: No hyphenation pattern has been}%
\typeout{** loaded for the language `#1'. Using the pattern for}%
\typeout{** the default language instead.}%
\else
\language=\csname l@#1\endcsname
\fi
#2}}
\providecommand{\BIBdecl}{\relax}
\BIBdecl

\bibitem{Collet2015}
A.~Collet, M.~Chuang, P.~Sweeney, D.~Gillett, D.~Evseev, D.~Calabrese,
  H.~Hoppe, A.~Kirk, and S.~Sullivan, ``High-quality streamable free-viewpoint
  video,'' \emph{ACM Transactions on Graphics (TOG)}, vol.~34, no.~4, p.~69,
  2015.

\bibitem{orts2016holoportation}
S.~Orts-Escolano, C.~Rhemann, S.~Fanello, W.~Chang, A.~Kowdle, Y.~Degtyarev,
  D.~Kim, P.~L. Davidson, S.~Khamis, M.~Dou \emph{et~al.}, ``Holoportation:
  Virtual 3d teleportation in real-time,'' in \emph{Proceedings of the 29th
  Annual Symposium on User Interface Software and Technology}.\hskip 1em plus
  0.5em minus 0.4em\relax ACM, 2016, pp. 741--754.

\bibitem{alexiadis2017}
D.~S. Alexiadis, A.~Chatzitofis, N.~Zioulis, O.~Zoidi, G.~Louizis, D.~Zarpalas,
  and P.~Daras, ``An integrated platform for live 3d human reconstruction and
  motion capturing,'' \emph{IEEE Transactions on Circuits and Systems for Video
  Technology}, vol.~27, no.~4, pp. 798--813, April 2017.

\bibitem{Robertini2017}
\BIBentryALTinterwordspacing
N.~Robertini, D.~Casas, E.~De~Aguiar, and C.~Theobalt, ``Multi-view performance
  capture of surface details,'' \emph{International Journal of Computer
  Vision}, vol. 124, no.~1, pp. 96--113, Aug 2017. [Online]. Available:
  \url{https://doi.org/10.1007/s11263-016-0979-1}
\BIBentrySTDinterwordspacing

\bibitem{Theobalt2013}
G.~Ye, Y.~Liu, Y.~Deng, N.~Hasler, X.~Ji, Q.~Dai, and C.~Theobalt,
  ``Free-viewpoint video of human actors using multiple handheld kinects,''
  \emph{IEEE Transactions on Cybernetics}, vol.~43, no.~5, pp. 1370--1382, Oct
  2013.

\bibitem{de2008performance}
E.~De~Aguiar, C.~Stoll, C.~Theobalt, N.~Ahmed, H.-P. Seidel, and S.~Thrun,
  ``Performance capture from sparse multi-view video,'' \emph{ACM Transactions
  on Graphics (TOG)}, vol.~27, no.~3, p.~98, 2008.

\bibitem{gall2009motion}
J.~Gall, C.~Stoll, E.~De~Aguiar, C.~Theobalt, B.~Rosenhahn, and H.-P. Seidel,
  ``Motion capture using joint skeleton tracking and surface estimation,'' in
  \emph{Computer Vision and Pattern Recognition, 2009. CVPR 2009. IEEE
  Conference on}.\hskip 1em plus 0.5em minus 0.4em\relax IEEE, 2009, pp.
  1746--1753.

\bibitem{beck2017sweeping}
S.~Beck and B.~Froehlich, ``Sweeping-based volumetric calibration and
  registration of multiple rgbd-sensors for 3d capturing systems,'' in
  \emph{Virtual Reality (VR), 2017 IEEE}.\hskip 1em plus 0.5em minus
  0.4em\relax IEEE, 2017, pp. 167--176.

\bibitem{kubota2007multiview}
A.~Kubota, A.~Smolic, M.~Magnor, M.~Tanimoto, T.~Chen, and C.~Zhang,
  ``Multiview imaging and 3dtv,'' \emph{IEEE signal processing magazine},
  vol.~24, no.~6, pp. 10--21, 2007.

\bibitem{kanade1997virtualized}
T.~Kanade, P.~Rander, and P.~Narayanan, ``Virtualized reality: Constructing
  virtual worlds from real scenes,'' \emph{IEEE multimedia}, vol.~4, no.~1, pp.
  34--47, 1997.

\bibitem{kauff2002immersive}
P.~Kauff and O.~Schreer, ``An immersive 3d video-conferencing system using
  shared virtual team user environments,'' in \emph{Proceedings of the 4th
  international conference on Collaborative virtual environments}.\hskip 1em
  plus 0.5em minus 0.4em\relax ACM, 2002, pp. 105--112.

\bibitem{joo2017panoptic}
H.~Joo, T.~Simon, X.~Li, H.~Liu, L.~Tan, L.~Gui, S.~Banerjee, T.~S. Godisart,
  B.~Nabbe, I.~Matthews \emph{et~al.}, ``Panoptic studio: A massively multiview
  system for social interaction capture,'' \emph{IEEE Transactions on Pattern
  Analysis and Machine Intelligence}, 2017.

\bibitem{lou2005real}
J.-G. Lou, H.~Cai, and J.~Li, ``A real-time interactive multi-view video
  system,'' in \emph{Proceedings of the 13th annual ACM international
  conference on Multimedia}.\hskip 1em plus 0.5em minus 0.4em\relax ACM, 2005,
  pp. 161--170.

\bibitem{marton2011real}
F.~Marton, E.~Gobbetti, F.~Bettio, J.~A.~I. Guiti{\'a}n, and R.~Pintus, ``A
  real-time coarse-to-fine multiview capture system for all-in-focus rendering
  on a light-field display,'' in \emph{3DTV Conference: The True
  Vision-Capture, Transmission and Display of 3D Video (3DTV-CON), 2011}.\hskip
  1em plus 0.5em minus 0.4em\relax IEEE, 2011, pp. 1--4.

\bibitem{tsai1987versatile}
R.~Tsai, ``A versatile camera calibration technique for high-accuracy 3d
  machine vision metrology using off-the-shelf tv cameras and lenses,''
  \emph{IEEE Journal on Robotics and Automation}, vol.~3, no.~4, pp. 323--344,
  1987.

\bibitem{kim2008design}
Y.~M. Kim, D.~Chan, C.~Theobalt, and S.~Thrun, ``Design and calibration of a
  multi-view tof sensor fusion system,'' in \emph{Computer Vision and Pattern
  Recognition Workshops, 2008. CVPRW'08. IEEE Computer Society Conference
  on}.\hskip 1em plus 0.5em minus 0.4em\relax IEEE, 2008, pp. 1--7.

\bibitem{ahmed2014using}
N.~Ahmed and I.~Junejo, ``Using multiple rgb-d cameras for 3d video acquisition
  and spatio-temporally coherent 3d animation reconstruction,''
  \emph{International Journal of Computer Theory and Engineering}, vol.~6,
  no.~6, p. 447, 2014.

\bibitem{berger2011markerless}
K.~Berger, K.~Ruhl, Y.~Schroeder, C.~Bruemmer, A.~Scholz, and M.~A. Magnor,
  ``Markerless motion capture using multiple color-depth sensors.'' in
  \emph{VMV}, 2011, pp. 317--324.

\bibitem{omnikinect2012}
\BIBentryALTinterwordspacing
B.~Kainz, S.~Hauswiesner, G.~Reitmayr, M.~Steinberger, R.~Grasset, L.~Gruber,
  E.~Veas, D.~Kalkofen, H.~Seichter, and D.~Schmalstieg, ``Omnikinect:
  Real-time dense volumetric data acquisition and applications,'' in
  \emph{Proceedings of the 18th ACM Symposium on Virtual Reality Software and
  Technology}, ser. VRST '12.\hskip 1em plus 0.5em minus 0.4em\relax New York,
  NY, USA: ACM, 2012, pp. 25--32. [Online]. Available:
  \url{http://doi.acm.org/10.1145/2407336.2407342}
\BIBentrySTDinterwordspacing

\bibitem{Zioulis2016}
N.~Zioulis, D.~Alexiadis, A.~Doumanoglou, G.~Louizis, K.~Apostolakis,
  D.~Zarpalas, and P.~Daras, ``3d tele-immersion platform for interactive
  immersive experiences between remote users,'' in \emph{2016 IEEE
  International Conference on Image Processing (ICIP)}, Sept 2016, pp.
  365--369.

\bibitem{kowalski2015live}
M.~Kowalski, J.~Naruniec, and M.~Daniluk, ``Live scan3d: A fast and inexpensive
  3d data acquisition system for multiple kinect v2 sensors,'' in \emph{3D
  Vision (3DV), 2015 International Conference on}.\hskip 1em plus 0.5em minus
  0.4em\relax IEEE, 2015, pp. 318--325.

\bibitem{zhang00}
Z.~Zhang, ``A flexible new technique for camera calibration,'' \emph{IEEE
  Transactions on Pattern Analysis and Machine Intelligence}, vol.~22, no.~11,
  pp. 1330--1334, Nov 2000.

\bibitem{fornaser2017automatic}
A.~Fornaser, P.~Tomasin, M.~De~Cecco, M.~Tavernini, and M.~Zanetti, ``Automatic
  graph based spatiotemporal extrinsic calibration of multiple kinect v2 tof
  cameras,'' \emph{Robotics and Autonomous Systems}, vol.~98, pp. 105--125,
  2017.

\bibitem{su2018fast}
P.-C. Su, J.~Shen, W.~Xu, S.-C.~S. Cheung, and Y.~Luo, ``A fast and robust
  extrinsic calibration for rgb-d camera networks,'' \emph{Sensors}, vol.~18,
  no.~1, p. 235, 2018.

\bibitem{wscg2018}
\BIBentryALTinterwordspacing
A.~Papachristou, N.~Zioulis, D.~Zarpalas, and P.~Daras, ``Markerless
  structure-based multi-sensor calibration for free viewpoint video capture,''
  in \emph{Proceedings of 26th International Conference in Central Europe on
  Computer Graphics, Visualization and Computer Vision'2018}, ser. WSCG '18,
  2018, pp. 88--97. [Online]. Available:
  \url{http://wscg.zcu.cz/WSCG2018/!!_CSRN-2801.pdf}
\BIBentrySTDinterwordspacing

\bibitem{keselmanintel}
L.~Keselman, J.~Iselin~Woodfill, A.~Grunnet-Jepsen, A.~Bhowmik, M.~Gupta,
  A.~Jauhari, K.~Kulkarni, S.~Jayasuriya, A.~Molnar, P.~Turaga \emph{et~al.},
  ``Intel realsense stereoscopic depth cameras,'' in \emph{The IEEE Conference
  on Computer Vision and Pattern Recognition (CVPR) Workshops}.

\bibitem{turbojpeg}
``{TurboJPEG},'' https://github.com/libjpeg-turbo/libjpeg-turbo, accessed:
  2018-09-03.

\bibitem{blosc}
``{Blosc},'' https://github.com/Blosc/c-blosc, accessed: 2018-09-03.

\bibitem{livescan3d}
M.~Kowalski, J.~Naruniec, and M.~Daniluk, ``Livescan3d: A fast and inexpensive
  3d data acquisition system for multiple kinect v2 sensors,'' in \emph{2015
  International Conference on 3D Vision}, Oct 2015, pp. 318--325.

\bibitem{kendall2015posenet}
A.~Kendall, M.~Grimes, and R.~Cipolla, ``Posenet: A convolutional network for
  real-time 6-dof camera relocalization,'' in \emph{Proceedings of the IEEE
  international conference on computer vision}, 2015, pp. 2938--2946.

\bibitem{krahenbuhl2011efficient}
P.~Kr{\"a}henb{\"u}hl and V.~Koltun, ``Efficient inference in fully connected
  crfs with gaussian edge potentials,'' in \emph{Advances in neural information
  processing systems}, 2011, pp. 109--117.

\bibitem{kendall1989survey}
D.~G. Kendall, ``A survey of the statistical theory of shape,''
  \emph{Statistical Science}, pp. 87--99, 1989.

\bibitem{jia2014caffe}
Y.~Jia, E.~Shelhamer, J.~Donahue, S.~Karayev, J.~Long, R.~Girshick,
  S.~Guadarrama, and T.~Darrell, ``Caffe: Convolutional architecture for fast
  feature embedding,'' \emph{arXiv preprint arXiv:1408.5093}, 2014.

\bibitem{kingma2014adam}
D.~P. Kingma and J.~Ba, ``Adam: A method for stochastic optimization,''
  \emph{arXiv preprint arXiv:1412.6980}, 2014.

\end{thebibliography}

\end{document}